\documentclass[11pt,a4paper]{article}
\usepackage{authblk}
\usepackage[hyperref]{acl2021}
\usepackage{times}
\usepackage{latexsym}

\usepackage[T1]{fontenc}
\usepackage{amsmath}
\usepackage{booktabs}
\usepackage{graphicx, accents}
\usepackage{standalone}
\usepackage{subcaption}
\usepackage{bm}
\usepackage{multirow}
\usepackage{float}
\usepackage{soul}
\usepackage{color}
\usepackage{siunitx}

\definecolor{light-gray}{gray}{0.85}
\definecolor{dark-gray}{gray}{0.55}

\DeclareRobustCommand{\best}[1]{{\sethlcolor{dark-gray}\hl{#1}}}
\DeclareRobustCommand{\second}[1]{{\sethlcolor{light-gray}\hl{#1}}}

\usepackage{tikz}
\usetikzlibrary{shapes,arrows,decorations.pathmorphing,backgrounds,positioning,fit,petri,calc}
\usepackage{tikzscale}

  \DeclareMathOperator{\BERT}{BERT}
  
\usepackage{microtype}

\aclfinalcopy 

\title{Superbizarre Is Not Superb:
Derivational Morphology Improves\\
BERT's Interpretation
of Complex Words}

\author[*$\ddag$]{Valentin Hofmann}
\author[$\dag$*]{Janet B. Pierrehumbert}
\author[$\ddag$]{Hinrich Sch\"utze}

\affil[*]{Faculty of Linguistics, University of Oxford}
\affil[$\dag$]{Department of Engineering Science, University of Oxford}
\affil[$\ddag$]{Center for Information and Language Processing, LMU Munich \protect\\ \texttt{valentin.hofmann@ling-phil.ox.ac.uk}}

\date{}

\begin{document}
\maketitle
\begin{abstract}
How does the input segmentation of pretrained language models (PLMs) 
affect their interpretations of complex words?
We present the first study investigating this question,
taking BERT as the example PLM and focusing on its semantic representations
of English derivatives.
We show that PLMs can be interpreted as serial dual-route models, i.e.,
the meanings of complex words are either stored or else need to 
be computed from the subwords,
which implies that maximally meaningful input tokens should
allow for the best generalization on new words. This hypothesis
is confirmed by a series of semantic probing 
tasks on which DelBERT (Derivation leveraging BERT), a model with 
derivational input segmentation, substantially outperforms
BERT with WordPiece segmentation. Our results suggest
 that the generalization capabilities of PLMs
could be further improved if a morphologically-informed vocabulary 
of input tokens were used.
\end{abstract}

\section{Introduction}

Pretrained language models (PLMs)
such as BERT \citep{Devlin.2019}, GPT-2 \citep{Radford.2019}, XLNet \citep{Yang.2019}, ELECTRA \citep{Clark.2020}, and T5 \citep{Raffel.2020} have yielded substantial improvements on a range of NLP tasks.
What linguistic properties do they have?
Various studies have tried to illuminate this question, with a focus on syntax \citep{Hewitt.2019, Jawahar.2019} and semantics \citep{Ethayarajh.2019,Ettinger.2020, Vulic.2020}.

One common characteristic of PLMs is 
their input segmentation: PLMs are based on fixed-size 
vocabularies of words and subwords
that are generated by compression algorithms
such as byte-pair encoding \citep{Gage.1994, Sennrich.2016} and 
WordPiece \citep{Schuster.2012, Wu.2016}. The segmentations
produced by these algorithms 
are linguistically questionable at times \citep{Church.2020},
which has been shown to worsen performance on certain
downstream tasks \citep{Bostrom.2020, Hofmann.2020c}. 
However, the wider implications of these
findings, particularly with regard to the generalization capabilities of PLMs, are
still 
poorly understood.

Here, we address a central aspect of this issue, 
namely how the input segmentation affects the semantic 
representations of PLMs, taking BERT as the example PLM. We focus on derivationally
complex words such as \texttt{superbizarre} since they exhibit systematic
patterns on the lexical level, providing an ideal testbed for linguistic 
generalization.
At the same time, the fact that low-frequency 
and out-of-vocabulary words are often
derivationally complex \citep{Baayen.1991} makes our work relevant in practical settings,
especially 
when many one-word expressions are involved, e.g., 
in query processing \citep{Kacprzak.2017}.

\begin{figure}
        \centering
        \begin{subfigure}[b]{0.23\textwidth}
            \centering
            \begin{tikzpicture}
[thick,
 model/.style={draw,rectangle,inner sep=4pt,rounded corners=6pt},
 element/.style={draw,rectangle,inner sep=2pt,rounded corners=6pt},
 token/.style={draw,rectangle,inner sep=0pt,rounded corners=4pt,
 minimum width={width("bizarre")+14pt},minimum height=0.75cm,text height=1.5ex,
 text depth=.25ex,font=\ttfamily},
 input/.style={draw,tape,tape bend top=none,inner sep=0pt,
 minimum height=0.75cm,font=\ttfamily},
 inputx/.style={draw=none,font=\ttfamily,inner sep=0pt,minimum width=25mm,minimum height=5mm,
 text height=1.5ex,text depth=.25ex,text width=23mm},
 inputy/.style={draw=none,font=\ttfamily,inner sep=0pt,minimum width=8mm,minimum height=5mm,
 text height=1.5ex,text depth=.25ex,text width=6mm},
 comp/.style={draw,rectangle,fill=white,inner sep=0pt,minimum width=5cm,minimum height=0.75cm},
 result/.style={draw=none,rectangle,fill=white,inner sep=2pt},
]

	
\node [inputx] (ix-0) at (0,0) {\small superbizarre};
\node [inputy, right=0.3 of ix-0] (iy-0) {\small neg};
\node [inputx, below=0.01 of ix-0] (ix-1) {\small applausive};
\node [inputy, below=0.01 of iy-0] (iy-1) {\small pos};

\node [input,fit=(ix-0) (iy-1)] (input) {};

\node [token,above=0.5 of input] (tok-1) {\small \#\#iza};
\node [token,left=0.1 of tok-1] (tok-0) {\small superb};
\node [token,right=0.1 of tok-1] (tok-2) {\small \#\#rre};

\begin{scope}[on background layer]	
\node [model,fit=(tok-0) (tok-2),orange!50,fill=orange!5,very thick,
label={[xshift=2.75cm,yshift=0.05cm,orange!50]below:$s_w$}] (tokenization) {};
\node [element,fit=(ix-0),black!50,fill=black!5,very thick,
label={[xshift=0.05cm,yshift=-0.4cm,black!50]left:$x$}](inputx) {};
\node [element,fit=(iy-0),black!50,fill=black!5,very thick,
label={[xshift=-0.05cm,yshift=-0.4cm,black!50]right:$y$}] (inputy) {};
\end{scope}
%
%

\node [token,draw=none,fill=none,above=0.5 of tok-0] (c-0) {};
\node [token,draw=none,fill=none,above=0.5 of tok-1] (c-1) {};
\node [token,draw=none,fill=none,above=0.5 of tok-2] (c-2) {};

\node[comp,above=0.5 of tok-1](bert) {$\BERT$};
\node[result,above=0.5 of bert](prob) {$p(y | s_w(x)) = .149$};

\draw[->,dotted,very thick,black!50,rounded corners=5pt] (inputx.west) -- ++(-12mm,0) |- (tokenization.west)  ; 
\draw[->,dotted,very thick,black!50,rounded corners=5pt] (inputy.east) -- ++(12mm,0) -- ++ (0,4.32cm) -| ([xshift=-1cm]prob.north)  ; 
	
\foreach \x in {0,1,2}
	\draw [->] (tok-\x) -- (tok-\x |- bert.south);

\draw [->] (bert) -- (prob);

\end{tikzpicture}
  \caption[]%
            {{\small BERT ($s_w$)}}   
  \label{fig:bert-ws}
        \end{subfigure} \hspace{1mm}
        \begin{subfigure}[b]{0.23\textwidth}  
            \centering 
            \begin{tikzpicture}
[thick,
 model/.style={draw,rectangle,inner sep=4pt,rounded corners=6pt},
 element/.style={draw,rectangle,inner sep=2pt,rounded corners=6pt},
 token/.style={draw,rectangle,inner sep=0pt,rounded corners=4pt,
 minimum width={width("bizarre")+14pt},minimum height=0.75cm,text height=1.5ex,
 text depth=.25ex,font=\ttfamily},
 input/.style={draw,tape,tape bend top=none,inner sep=0pt,
 minimum height=0.75cm,font=\ttfamily},
 inputx/.style={draw=none,font=\ttfamily,inner sep=0pt,minimum width=25mm,minimum height=5mm,
 text height=1.5ex,text depth=.25ex,text width=23mm},
 inputy/.style={draw=none,font=\ttfamily,inner sep=0pt,minimum width=8mm,minimum height=5mm,
 text height=1.5ex,text depth=.25ex,text width=6mm},
 comp/.style={draw,rectangle,fill=white,inner sep=0pt,minimum width=5cm,minimum height=0.75cm},
 result/.style={draw=none,rectangle,fill=white,inner sep=2pt},
]

	
\node [inputx] (ix-0) at (0,0) {\small superbizarre};
\node [inputy, right=0.3 of ix-0] (iy-0) {\small neg};
\node [inputx, below=0.01 of ix-0] (ix-1) {\small applausive};
\node [inputy, below=0.01 of iy-0] (iy-1) {\small pos};

\node [input,fit=(ix-0) (iy-1)] (input) {};

\node [token,above=0.5 of input] (tok-1) {\small -};
\node [token,left=0.1 of tok-1] (tok-0) {\small super};
\node [token,right=0.1 of tok-1] (tok-2) {\small bizarre};

\begin{scope}[on background layer]	
\node [model,fit=(tok-0) (tok-2),blue!50,fill=blue!5,very thick,
label={[xshift=2.75cm,yshift=0.05cm,blue!50]below:$s_d$}] (tokenization) {};
\node [element,fit=(ix-0),black!50,fill=black!5,very thick,
label={[xshift=0.05cm,yshift=-0.4cm,black!50]left:$x$}](inputx) {};
\node [element,fit=(iy-0),black!50,fill=black!5,very thick,
label={[xshift=-0.05cm,yshift=-0.4cm,black!50]right:$y$}] (inputy) {};
\end{scope}
%
%

\node [token,draw=none,fill=none,above=0.5 of tok-0] (c-0) {};
\node [token,draw=none,fill=none,above=0.5 of tok-1] (c-1) {};
\node [token,draw=none,fill=none,above=0.5 of tok-2] (c-2) {};

\node[comp,above=0.5 of tok-1](bert) {$\BERT$};
\node[result,above=0.5 of bert](prob) {$p(y | s_d(x)) = .931$};

\draw[->,dotted,very thick,black!50,rounded corners=5pt] (inputx.west) -- ++(-12mm,0) |- (tokenization.west)  ; 
\draw[->,dotted,very thick,black!50,rounded corners=5pt] (inputy.east) -- ++(12mm,0) -- ++ (0,4.32cm) -| ([xshift=-1cm]prob.north)  ; 
	
\foreach \x in {0,1,2}
	\draw [->] (tok-\x) -- (tok-\x |- bert.south);

\draw [->] (bert) -- (prob);

\end{tikzpicture}%
            \caption[]%
            {{\small DelBERT ($s_d$)}}    
            \label{fig:bert-ds}
        \end{subfigure}
        \caption[]{Basic experimental setup. BERT with WordPiece
          segmentation ($s_w$) mixes part of the stem
          \texttt{bizarre} with the prefix \texttt{super}, creating an
          association with \texttt{superb} (left panel). DelBERT with derivational
          segmentation ($s_d$), on the other hand, separates
          prefix and stem by a hyphen (right panel). The two
          likelihoods
          are averaged across 20 models trained with
          different random seeds. The average likelihood of the true class is considerably higher with DelBERT than with BERT. While \texttt{superbizarre}
          has negative sentiment, \texttt{applausive} is an example
          of a complex word with positive sentiment.
        }
        \label{fig:model}
\end{figure}
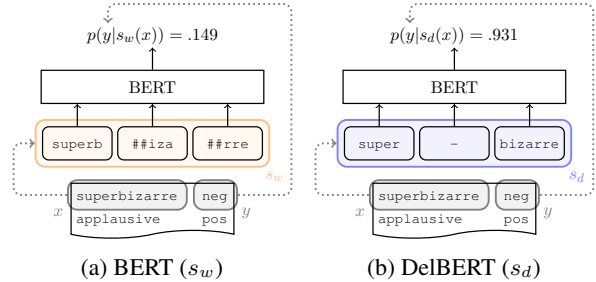

The topic of this paper is related to the more fundamental
question of how
PLMs represent the meaning of complex words in the
first place. So far, most studies have focused 
on methods of representation extraction,
using ad-hoc heuristics such as averaging the subword 
embeddings \citep{Pinter.2020b,
Sia.2020, Vulic.2020}
or taking the first subword embedding \citep{Devlin.2019, Heinzerling.2019, Martin.2020}.
While not resolving the issue, we lay the theoretical groundwork for more 
systematic analyses
by showing that PLMs can be regarded as serial dual-route models \citep{Caramazza.1988}, i.e.,
the meanings of complex words are either stored or else need to 
be computed from the subwords.

\textbf{Contributions.} We present the first study 
examining how the input segmentation of PLMs, specifically BERT, 
affects their interpretations of derivationally complex English words. We show that PLMs can be interpreted as serial dual-route models,
which implies that maximally meaningful input tokens should
allow for the best generalization on new words. This hypothesis
is confirmed by a series of semantic probing 
tasks on which derivational segmentation substantially outperforms
BERT's WordPiece segmentation. This
suggests that the generalization capabilities of PLMs
could be further improved if a morphologically-informed vocabulary 
of input tokens were used. We also 
publish three large datasets 
of derivationally complex words with corresponding semantic 
properties.\footnote{We make our code and data available at \url{https://github.com/valentinhofmann/superbizarre}.}

\section{How Are Complex Words Processed?}

\subsection{Complex Words in Psycholinguistics}

The question of how complex words are processed has been
at the center of psycholinguistic research over the 
last decades (see \citet{Leminen.2019} for a recent review).
Two basic processing mechanisms have been proposed: \textit{storage}, where 
the meaning of
complex words is listed in the mental lexicon 
\citep{Manelis.1977, Butterworth.1983b, Feldman.1987, Bybee.1988, Stemberger.1994, Bybee.1995, Bertram.2000b},
and \textit{computation},
where the meaning of complex words 
is inferred based on the meaning
of stem and affixes
\citep{Taft.1975, Taft.1979, Taft.1981, Taft.1988,  Taft.1991, Taft.1994, Rastle.2004, Taft.2004, Rastle.2008}. 

In contrasting with \textit{single-route} frameworks,  \textit{dual-route} models  allow
for a combination of storage and computation. Dual-route models are further classified by whether  
they regard the processes of retrieving meaning from the mental lexicon and computing meaning based on stem and affixes as \textit{parallel}, i.e., both mechanisms are always activated \citep{Frauenfelder.1992, Schreuder.1995, Baayen.1997, Baayen.2000, 
Bertram.2000c, New.2004, Kuperman.2008, Kuperman.2009}, or \textit{serial}, i.e., the computation-based mechanism is only activated when
the storage-based one fails \citep{Laudanna.1985, Burani.1987,  Caramazza.1988, Burani.1992, Laudanna.1995, Alegre.1999b}.

Outside the taxonomy presented so far are recent models that assume multiple levels of representation as well as various forms of interaction between them \citep{Racz.2015, Needle.2018}. In these models,
 sufficiently frequent complex words are stored together with representations that include their internal structure. Complex-word processing is driven by analogical processes over the mental lexicon \citep{Racz.2020}.

\subsection{Complex Words in NLP and PLMs} \label{sec:complex-nlp}

Most models of word meaning proposed in NLP can be roughly
assigned to either the single-route or dual-route approach. Word embeddings that represent complex words as whole-word vectors \citep{Deerwester.1990, Mikolov.2013b, Mikolov.2013c, Pennington.2014} can be seen as single-route storage models. Word embeddings
that represent complex words as a function of subword or morpheme vectors \citep{Schutze.1992b,Luong.2013}
can be seen as single-route computation models. Finally, word embeddings
that represent complex words as a function of subword or morpheme vectors as well 
as whole-word vectors \citep{Botha.2014, Qiu.2014, Bhatia.2016, Bojanowski.2017, Athiwaratkun.2018, Salle.2018} are most closely related to parallel dual-route approaches.

Where are PLMs to be located
in this taxonomy? PLMs represent many complex words as whole-word vectors (which are fully stored). 
Similarly to how character-based models represent word meaning \citep{Kim.2016, Adel.2017}, they can also store
the meaning of frequent complex words that are segmented into subwords, i.e., frequent 
subword collocations, in their model weights. When 
the complex-word meaning is neither stored as a whole-word vector nor in the model weights, 
PLMs compute the meaning as a compositional function of the subwords. Conceptually, PLMs can thus be interpreted as serial dual-route models.
While the parallelism has
not been observed before, it follows logically from the structure of PLMs. 
The key goal of this paper is to show that the implications
of this observation
are borne out empirically.

As a concrete example, consider the 
complex words \texttt{stabilize}, \texttt{realize}, \texttt{finalize}, \texttt{mobilize},
\texttt{tribalize}, and \texttt{templatize}, which are all formed by adding the verbal suffix \texttt{ize} 
to a nominal or adjectival stem. Taking BERT, specifically BERT\textsubscript{BASE} (uncased) \citep{Devlin.2019}, as the example PLM, the words \texttt{stabilize} and \texttt{realize}
have individual tokens in the input vocabulary and are hence associated 
with whole-word vectors storing their meanings, including highly lexicalized meanings as in the case of \texttt{realize}.
By contrast, the words \texttt{finalize} and \texttt{mobilize} are segmented into 
\texttt{final}, \texttt{\#\#ize} and \texttt{mob}, \texttt{\#\#ili}, \texttt{\#\#ze}, 
which entails that their meanings are not stored as whole-word vectors. However, both words 
have relatively high absolute frequencies of 2,540 (\texttt{finalize}) and 6,904 (\texttt{mobilize})
in the English Wikipedia, the main dataset used to pretrain BERT \citep{Devlin.2019}, which means 
that BERT can store their meanings in its model weights during pretraining.\footnote{Previous research suggests that such lexical 
knowledge is stored in the lower layers of BERT \citep{Vulic.2020}.} Notice this is even possible
in the case of highly lexicalized meanings as for \texttt{mobilize}. Finally, the words \texttt{tribalize} and \texttt{templatize}
are segmented into \texttt{tribal}, \texttt{\#\#ize} and \texttt{te}, \texttt{\#\#mp}, \texttt{\#\#lat}, \texttt{\#\#ize}, but 
as opposed to \texttt{finalize} and \texttt{mobilize} they do not occur in the English Wikipedia.
As a result, BERT cannot store their meanings in its model weights during pretraining and needs to compute them from 
the meanings of the subwords.

Seeing PLMs as serial dual-route models allows for a more nuanced view on the central research question 
of this paper: in order to investigate semantic generalization we need to
investigate the representations of those complex words that activate the
computation-based route. The words that do so are the ones
whose  
meaning is 
neither stored as a whole-word vector nor in the model
weights and hence needs to be computed compositionally as a
function of the subwords (\texttt{tribalize} and \texttt{templatize} in the discussed examples).
 We hypothesize that the morphological validity of the segmentation affects the representational quality in these cases, and that the best generalization is achieved by maximally meaningful tokens. It is crucial to note this does not imply that the tokens have to be morphemes, but the 
segmentation boundaries need to
coincide with morphological boundaries, i.e., groups of morphemes (e.g., \texttt{tribal} in the segmentation of \texttt{tribalize}) are also possible.\footnote{This is in line with substantial evidence from linguistics showing that 
frequent groups of morphemes can be treated as semantic wholes \citep{Stump.2017, Stump.2019}.}
For \texttt{tribalize} and \texttt{templatize}, we therefore expect the segmentation \texttt{tribal}, \texttt{\#\#ize}
(morphologically valid since all segmentation boundaries are morpheme boundaries) to result in a representation of higher quality
than the segmentation \texttt{te}, \texttt{\#\#mp}, \texttt{\#\#lat}, \texttt{\#\#ize} (morphologically invalid since the boundaries between 
\texttt{te}, \texttt{\#\#mp}, and \texttt{\#\#lat} are not morpheme boundaries).
On the other hand, complex words
whose meanings are stored in the model weights (\texttt{finalize} and \texttt{mobilize} 
in the discussed examples) are expected to be 
affected by the segmentation to a much lesser 
extent: if the meaning of a complex word is stored in the model weights, it should matter
less
whether the specific segmentation activating that meaning is morphologically valid (\texttt{final}, \texttt{\#\#ize}) or not (\texttt{mob}, \texttt{\#\#ili}, \texttt{\#\#ze}).\footnote{We expect the distinction between storage and computation
of complex-word meaning for PLMs
to be a continuum. While the findings presented here are consistent with 
this view, we defer a more in-depth analysis to future work.}

\section{Experiments}

\begin{table*} [t!]\centering
\resizebox{\linewidth}{!}{%
\begin{tabular}{@{}llrllll@{}}
\toprule
{} & {} & {} & \multicolumn{2}{c}{Class 1} & \multicolumn{2}{c}{Class 2} \\
\cmidrule(lr){4-5}
\cmidrule(l){6-7}
Dataset & Dimension & $|\mathcal{D}|$ & Class & Examples & Class & Example \\
\midrule
Amazon & Sentiment & 239,727 & \texttt{neg} & \texttt{overpriced}, \texttt{crappy} & \texttt{pos} & \texttt{megafavorite}, \texttt{applausive}\\
ArXiv & Topicality & 97,410 & \texttt{phys} & \texttt{semithermal}, \texttt{ozoneless} & \texttt{cs} & \texttt{autoencoded}, \texttt{rankable} \\
Reddit & Topicality & 85,362  & \texttt{ent} & \texttt{supervampires}, \texttt{spoilerful} & \texttt{dis} & \texttt{antirussian}, \texttt{immigrationism}\\
\bottomrule
\end{tabular}}
\caption{Dataset characteristics. The table provides information about the datasets such as the 
relevant semantic dimensions with their classes and example
complex words. $|\mathcal{D}|$: number of complex words;
\texttt{neg}: negative;
\texttt{pos}: positive;
\texttt{phys}: physics;
\texttt{cs}: computer science;
\texttt{ent}: entertainment;
\texttt{dis}: discussion.
}  \label{tab:data-stats}
\end{table*}

\subsection{Setup}

Analyzing the impact of different segmentations on BERT's semantic
generalization capabilities is not straightforward
since it is not clear a priori how to measure the quality 
of representations. Here, we devise a novel 
lexical-semantic probing task: we use BERT's representations
for complex words to predict semantic dimensions, specifically sentiment and topicality 
(see Figure \ref{fig:model}). For sentiment, given
the example complex word  \texttt{superbizarre}, the task is to predict that 
its sentiment is negative. For topicality,
given the example complex word 
\texttt{isotopize}, the task is to predict that it is used in physics.
We confine ourselves 
to binary prediction, i.e., the probed semantic dimensions always consist of two classes (e.g., positive and negative). The extent to which a segmentation supports a solution of this task 
 is taken as an indicator
of its representational quality.

More formally, let $\mathcal{D}$ be a dataset consisting of complex words $x$ and corresponding classes $y$ that instantiate a certain semantic dimension (e.g., sentiment). We denote with $s(x) = (t_1, \dots, t_k)$ the segmentation of $x$ into a sequence of $k$ subwords. We ask how $s$ impacts the capability of BERT to predict $y$, i.e.,
how $p(y|(s(x))$, the likelihood of the true semantic class $y$ given a certain segmentation of $x$, depends on different choices for $s$. The two segmentation methods we compare in this study are BERT's standard WordPiece segmentation \citep{Schuster.2012, Wu.2016}, $s_w$, and a derivational segmentation that segments complex words into stems and affixes, $s_d$.

\subsection{Data} \label{sec:data}

\begin{table*} [t!]\centering
\resizebox{0.7\linewidth}{!}{%
\begin{tabular}{@{}lrrrrrr@{}}
\toprule
{} & \multicolumn{2}{c}{Amazon}  & \multicolumn{2}{c}{ArXiv} & \multicolumn{2}{c}{Reddit}   \\
\cmidrule(lr){2-3}
\cmidrule(lr){4-5}
\cmidrule(l){6-7}
Model & Dev & Test & Dev & Test  & Dev & Test \\
\midrule
DelBERT & \best{.635 $\pm$ .001} & \best{.639 $\pm$ .002} & \best{.731 $\pm$ .001} & \best{.723 $\pm$ .001} & \best{.696 $\pm$ .001} & \best{.701 $\pm$ .001} \\
BERT & \second{.619 $\pm$ .001} & \second{.624 $\pm$ .001} & .704 $\pm$ .001 & \second{.700 $\pm$ .002} & .664 $\pm$ .001 & .664 $\pm$ .003\\
\midrule
Stem & .572 $\pm$ .003 & .573 $\pm$ .003 & \second{.705 $\pm$ .001} & .697 $\pm$ .001 & \second{.679 $\pm$ .001} & \second{.684 $\pm$ .002}\\
Affixes & .536 $\pm$ .008 & .539 $\pm$ .008 & .605 $\pm$ .001 & .603 $\pm$ .002 & .596 $\pm$ .001 & .596 $\pm$ .001\\
\bottomrule
\end{tabular}
}
\caption{Results. The table shows the average performance as well as standard deviation (F1) of 20 models trained with different random seeds. Best result per column highlighted in gray, second-best in light gray.}  \label{tab:performance}
\end{table*}

\begin{figure*}[t!]
        \centering
        \includegraphics[width=\linewidth]{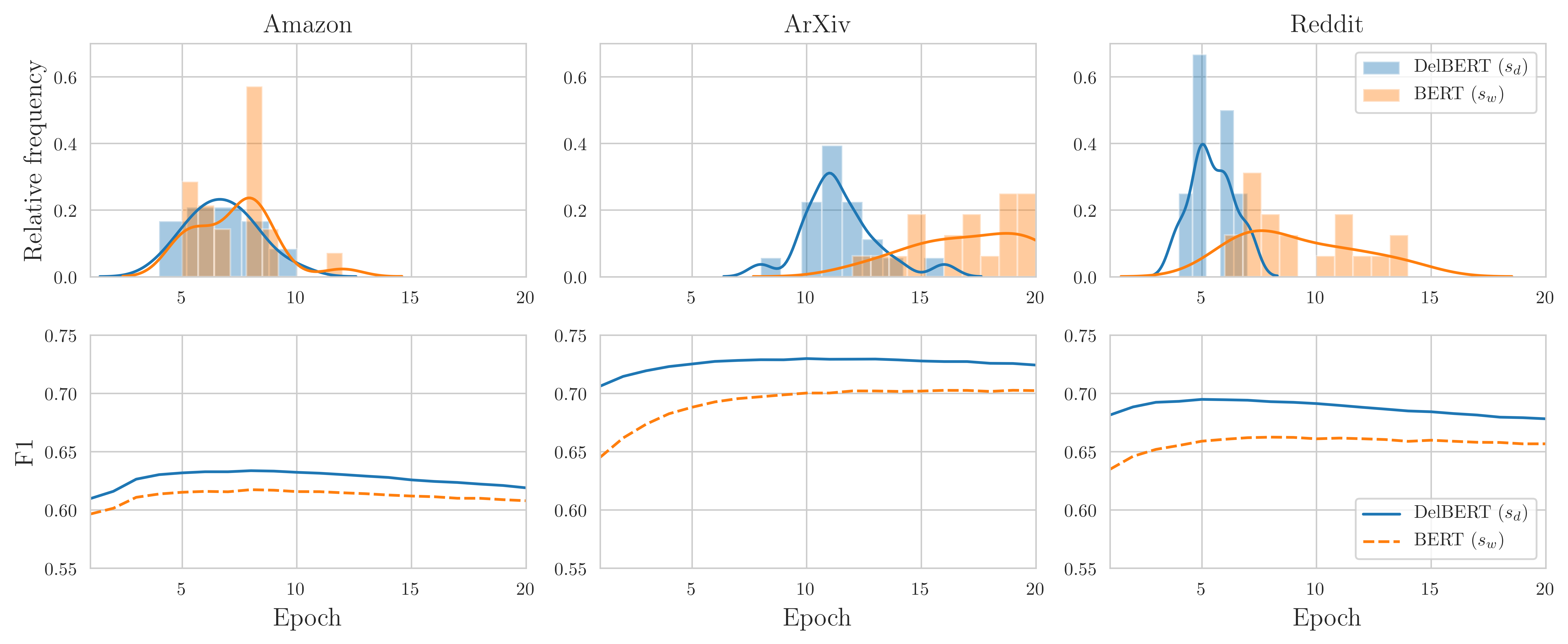}      
        \caption[]{Convergence analysis. The upper panels
          show the distributions of the number of epochs
          after which the models reach their maximum
          validation performance. The lower panels show the
          trajectories of the average validation performance (F1) across
          epochs. The plots are based on 20 models trained
          with different random seeds. The convergence statistics for DelBERT and BERT are directly comparable because the optimal learning rate is the same (see Appendix \ref{app:hyper}). DelBERT models reach their performance peak faster than 
          BERT models.}
        \label{fig:convergence}
\end{figure*}

Since existing datasets do not allow us to conduct experiments following the described setup, 
we create new datasets in a weakly-supervised fashion that is conceptually 
similar to the method proposed by \citet{Mintz.2009}:
we employ large datasets annotated for sentiment or topicality, extract derivationally complex words,
and use the dataset
labels to establish their semantic classes.

For determining and segmenting derivationally complex words, we use the algorithm introduced by \citet{Hofmann.2020a}, 
which takes as input a set of prefixes, suffixes, and stems and checks for each word in the data whether it can be derived from a stem using a combination of prefixes and suffixes.\footnote{The distinction between inflectionally and derivationally complex words is notoriously fuzzy \citep{Haspelmath.2010, Hacken.2014}. We try to exclude inflection as far as
possible (e.g., by removing problematic affixes such as \texttt{ing}) but are aware that a clear
separation does not exist.} The algorithm is
sensitive to morpho-orthographic rules of English
\cite{Plag.2003}, e.g., when the suffix \texttt{ize} is removed from
\texttt{isotopize}, the result is \texttt{isotope}, not
\texttt{isotop}. We follow \citet{Hofmann.2020c} in using the prefixes, suffixes, and stems in BERT's
WordPiece vocabulary as input to the algorithm. This means that all tokens used by the derivational segmentation
are in principle also available to the WordPiece segmentation, i.e., 
the difference between $s_w$ and $s_d$ does not lie in the vocabulary per se
but rather in the way the vocabulary is used. See Appendix \ref{app:segmentation} for details about the derivational segmentation.

To get the semantic classes, we compute for each complex word 
which fraction of texts containing the word belongs to one of 
two predefined sets of dataset labels (e.g., reviews with four and five stars for positive
sentiment) and rank
all words accordingly. We then take the first and third
tertiles of
complex words as representing the two classes. We randomly split
              the words into 60\% training, 20\% development,
              and 20\% test.

In the following, we describe the characteristics of the three datasets in greater depth.
Table \ref{tab:data-stats} provides summary statistics. See Appendix \ref{app:preprocessing} for details about data preprocessing.

\textbf{Amazon.} Amazon is an online e-commerce platform. A large 
dataset of Amazon reviews has been made publicly available \citep{Ni.2019}.\footnote{\url{https://nijianmo.github.io/amazon/index.html}} We extract derivationally complex words from reviews with one or two (\texttt{neg}) as well as four or five stars (\texttt{pos}), discarding three-star reviews
for a clearer separation \citep{Yang.2017}.

\textbf{ArXiv.} ArXiv is an open-access distribution service for scientific 
articles. Recently, a dataset of all papers published on ArXiv with associated metadata has been released.\footnote{\url{https://www.kaggle.com/Cornell-University/arxiv}} 
For this study, we extract all articles from physics (\texttt{phys}) and computer science (\texttt{cs}), which we identify using ArXiv's subject classification.
We choose physics and computer science since we expect large topical distances for these classes (compared to alternatives such as 
mathematics and computer science).

\textbf{Reddit.} Reddit is a social media
platform hosting discussions about various topics. It is divided into smaller communities, so-called subreddits, which have been shown to be a rich source of derivationally complex words \citep{Hofmann.2020b}. \citet{Hofmann.2020c} have published a dataset of derivatives found on Reddit annotated with
the subreddits in which they occur.\footnote{\url{https://github.com/valentinhofmann/dagobert}} Inspired
by a content-based subreddit categorization scheme,\footnote{\url{https://www.reddit.com/r/TheoryOfReddit/comments/1f7hqc/the_200_most_active_subreddits_categorized_by}}
we define two groups 
of subreddits, an entertainment set (\texttt{ent}) consisting of the subreddits \texttt{anime}, \texttt{DestinyTheGame}, \texttt{funny}, \texttt{Games}, \texttt{gaming},
\texttt{leagueoflegends}, \texttt{movies}, \texttt{Music}, \texttt{pics}, and \texttt{videos}, 
as well as a discussion set (\texttt{dis}) consisting of the 
subreddits \texttt{askscience}, \texttt{atheism}, \texttt{conspiracy}, \texttt{news},
\texttt{Libertarian}, \texttt{politics}, \texttt{science}, 
\texttt{technology}, \texttt{TwoXChromosomes}, and \texttt{worldnews}, and extract all derivationally complex words
occurring in them. We again expect large topical distances for these classes.

Given that the automatic creation of the datasets necessarily introduces noise,
we measure human 
performance on 100 randomly sampled words per dataset, which ranges between 71\% (Amazon) 
and 78\% (ArXiv). These values can thus be seen as an upper bound
on performance.

 \begin{figure*}[t!]
        \centering
        \includegraphics[width=\linewidth]{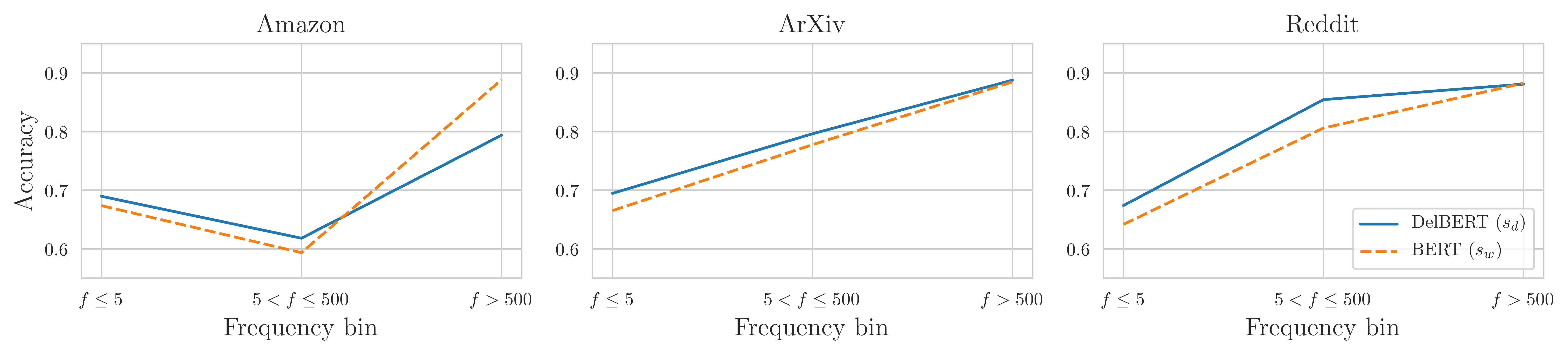}      
        \caption[]{Frequency analysis. The plots show the average performance (accuracy) of 20 BERT and DelBERT models trained with different random seeds for complex words of low ($f \leq 5$), mid ($5 < f \leq 500$), and high ($f > 500$) frequency. On all three datasets, BERT performs similarly or better than DelBERT for complex words of high frequency but worse for complex words of low and mid frequency.}
        \label{fig:frequency}
\end{figure*}

\subsection{Models}
We train two main models on each binary classification task: BERT with the standard
WordPiece segmentation ($s_w$) and BERT using the derivational segmentation ($s_d$), 
a model that we refer to as DelBERT (\textbf{De}rivation \textbf{l}everaging \textbf{BERT}).
BERT and DelBERT are identical except for the way in which they use the vocabulary of input tokens
(but the vocabulary itself is also identical for both models).
The specific BERT variant we use is
BERT\textsubscript{BASE} (uncased) \citep{Devlin.2019}.
For the derivational
segmentation, we follow previous work by \citet{Hofmann.2020c} 
in separating stem and prefixes by a hyphen.
We further follow \citet{Casanueva.2020} and \citet{Vulic.2020}
in mean-pooling the output representations for all subwords, excluding 
BERT's special tokens. The mean-pooled
representation is then fed into a two-layer feed-forward network 
for classification. To examine the relative importance of different types of morphological units,
we train two additional models in which we ablate information about  stems and
affixes, i.e., we represent stems and affixes by the same randomly chosen
input embedding.\footnote{For affix ablation, we use two different 
input embeddings for prefixes and suffixes.}

We finetune BERT, DelBERT, and the two ablated models on the three datasets using 20 different 
random seeds. We choose F1 as the evaluation measure. See Appendix \ref{app:hyper} for details about implementation and hyperparameters.

\subsection{Results} \label{sec-results}

DelBERT ($s_d$) outperforms BERT ($s_w$) by a large margin 
on all three datasets (Table \ref{tab:performance}). It is interesting to notice that the performance
difference is larger for ArXiv and Reddit than for Amazon, indicating that 
the gains in representational quality are particularly large for topicality.

What is it that leads to DelBERT's increased performance? The ablation study shows that
models using only stem information already achieve relatively high performance and are on par or even better than
the BERT models on ArXiv and Reddit.
However, the DelBERT models still perform substantially 
better than the stem models on all three datasets. The gap is particularly pronounced
for Amazon, which indicates that the interaction between the meaning of stem and affixes
is more complex for sentiment than for topicality. This makes sense from a linguistic point of view:
while stems tend to be good cues for the topical associations of a complex word, sentiment often depends on semantic interactions between stems and affixes.
For example, while the prefix \texttt{un} turns the
sentiment of \texttt{amusing} negative, it turns the sentiment of \texttt{biased} positive.
Such effects involving negation and antonymy are known to be challenging
for PLMs \citep{Ettinger.2020, Kassner.2020} and might be one of the reasons for the generally lower performance on Amazon.\footnote{Another reason for the lower performance on sentiment is that the datasets were created automatically (see Section~\ref{sec:data}), and hence
many complex words do not directly carry information about sentiment or topicality. The density of such words is higher for sentiment than topicality since the topic of discussion affects the likelihoods of most content words.}
The performance of models using only affixes is much lower.

\subsection{Quantitative Analysis} \label{sec:quant-analysis}

\begin{figure*}
        \centering
        \begin{subfigure}[h]{\textwidth}
            \centering
        \includegraphics[width=\textwidth]{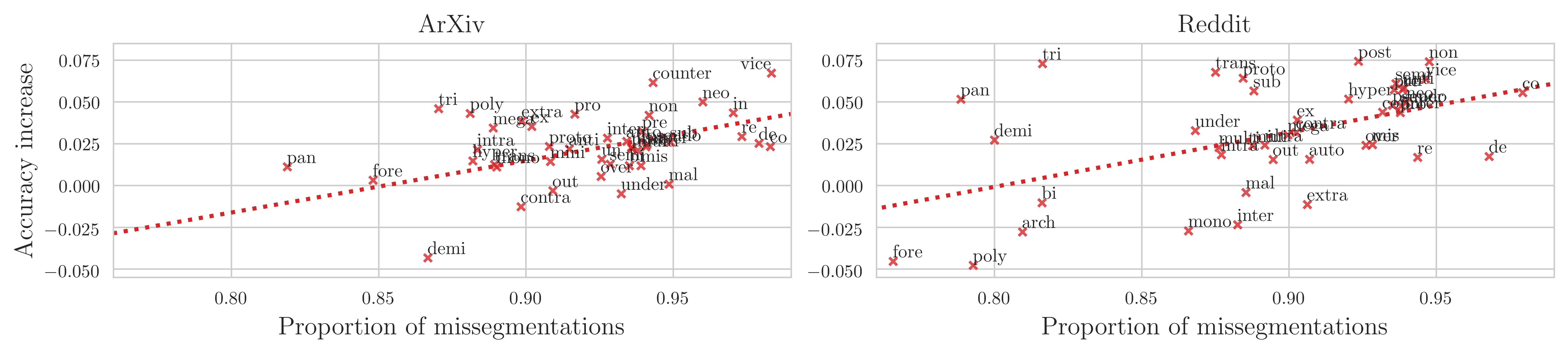}
  \caption[]%
            {{Topicality prediction}}   
  \label{fig:results-top}
                      \vspace{0.3cm}    
        \end{subfigure}
        \begin{subfigure}[h]{\textwidth}  
            \centering 
        \includegraphics[width=\textwidth]{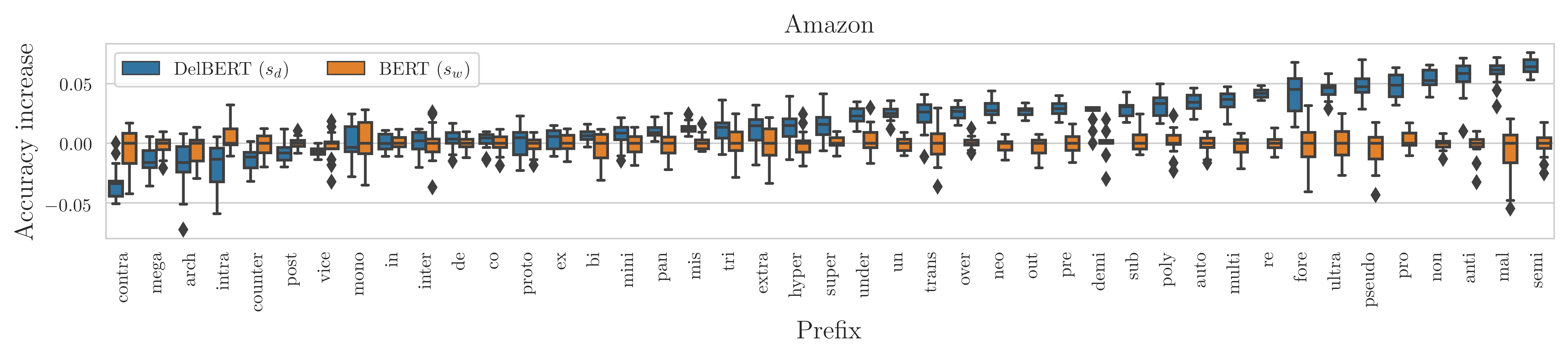}  
            \caption[]%
            {{Sentiment prediction}}    
            \label{fig:results-sent}
        \end{subfigure}
      
        \caption[]{Accuracy increase of DelBERT compared to BERT for prefixes. The plots show 
        the accuracy increase as a function of the proportion of morphologically incorrect WordPiece segmentations (topicality prediction) and as ordered boxplot pairs centered on the median accuracy of BERT (sentiment prediction). Negative values mean that 
        the DelBERT models have a lower accuracy than the BERT models for a certain prefix.}
        \label{fig:prefixes}
\end{figure*}

To further examine how BERT ($s_w$) and DelBERT ($s_d$) differ in the 
way they infer the meaning of complex words, 
we perform a convergence analysis. We find that
the DelBERT models reach their peak in performance faster than
the BERT models (Figure \ref{fig:convergence}). 
This is in line with our interpretation
of PLMs as serial
dual-route models (see Section \ref{sec:complex-nlp}): while
DelBERT operates on morphological units and can combine the subword meanings to infer the meanings of complex words, BERT's subwords do not necessarily carry lexical meanings, and hence the derivational patterns need
to be stored by adapting the model weights. This is an additional burden, leading
to longer convergence times and substantially worse overall performance.

Our hypothesis that PLMs can use two routes to process complex words
(storage in weights and compositional computation based on input embeddings), and that 
the second route is blocked when the input segmentation is not morphological, suggests
the existence of frequency effects: BERT might have seen frequent complex words
multiple times during pretraining and stored their meaning in the model weights. This
is less likely for infrequent complex words, making the capability to compositionally infer the meaning (i.e., the computation route) more important. We therefore expect the difference 
in performance between DelBERT (which should have
an advantage on the computation route) and BERT to be larger for infrequent words. To test this hypothesis,
we split the complex words of each dataset into three bins
of low ($f \leq 5$), mid ($5 < f \leq 500$), and high ($f >
500$) absolute frequencies, and analyze how the performance
of BERT and DelBERT differs on the three bins. For this and all subsequent analyses, we merge development and test sets 
and use accuracy instead of F1 since it makes comparisons across small sets of data points more interpretable. The results are in line with our hypothesis (Figure \ref{fig:frequency}): BERT performs worse than 
DelBERT on complex words of low and mid frequencies but achieves very similar (ArXiv, Reddit) or
even better (Amazon) accuracies on high-frequency complex words. These results
strongly suggest that two different mechanisms are involved, and that BERT 
has a disadvantage for complex words that do not have a high frequency. At the same time, the slight advantage of BERT on high-frequency complex words indicates that it has high-quality representations of these words in its weights, which DelBERT cannot exploit since 
it uses a different segmentation.

\begin{table*} [t!]\centering
\resizebox{\linewidth}{!}{%
\begin{tabular}{@{}llllrlr@{}}
\toprule
Dataset & $x$ & $y$ & $s_d(x)$ & $\mu_p$ & $s_w(x)$ & $\mu_p$  \\
\midrule
\multirow{3}{*}{Amazon} 
& \texttt{applausive} & \texttt{pos} & \texttt{applause, \#\#ive} & .847 & \texttt{app, \#\#laus, \#\#ive}& .029\\
& \texttt{superannoying} & \texttt{neg} & \texttt{super, -, annoying} & .967 & \texttt{super, \#\#ann, \#\#oy, \#\#ing} & .278\\
& \texttt{overseasoned} & \texttt{neg} & \texttt{over, -, seasoned} & .956 & \texttt{overseas, \#\#oned}
& .219\\
\midrule
\multirow{3}{*}{ArXiv}& \texttt{isotopize} & \texttt{phy} & \texttt{isotope, \#\#ize} & 
.985 & \texttt{iso, \#\#top, \#\#ize} & .039\\
& \texttt{antimicrosoft} & \texttt{cs} & \texttt{anti, -, microsoft} & .936 & \texttt{anti, \#\#mic, \#\#ros, \#\#oft} & .013\\
& \texttt{inkinetic} & \texttt{phy} & \texttt{in, -, kinetic} & .983 & \texttt{ink, \#\#ine, \#\#tic} &
.035\\
\midrule
\multirow{3}{*}{Reddit} & \texttt{prematuration} &\texttt{dis} & \texttt{premature, \#\#ation} & 
.848 & \texttt{prem, \#\#at, \#\#uration} & .089\\
 & \texttt{nonmultiplayer}& \texttt{ent} & \texttt{non, -, multiplayer} & .950 & \texttt{non, \#\#mu, \#\#lt, \#\#ip, \#\#layer} & .216\\
 & \texttt{promosque} & \texttt{dis} & \texttt{pro, -, mosque} & .961 & \texttt{promo, \#\#sque} & .066
\\
\bottomrule
\end{tabular}}
\caption{Error analysis. The table gives example complex
  words that are consistently classified correctly by
  DelBERT and incorrectly by BERT. $x$: complex word; $y$:
  semantic class; $s_d(x)$: derivational segmentation;
  $\mu_p$: average likelihood of true semantic class across
  20 models trained with different random seeds; $s_w(x)$:
  WordPiece segmentation. For the complex words shown, $\mu_p$ is considerably higher with DelBERT than with BERT.}  \label{tab:qual-analysis}
\end{table*}

We are further interested to see whether the affix type
has an impact on the relative performance of BERT and DelBERT. To examine this question, we measure the accuracy increase of DelBERT
as compared to BERT for individual affixes, averaged across datasets and 
random seeds. We find that the increase is almost twice as large for prefixes ($\mu = .023$, $\sigma= .017$) than for 
suffixes ($\mu = .013$, $\sigma= .016$), a difference that is shown to be significant 
by a two-tailed Welch's $t$-test ($d = .642$, $t(82.97) = 2.94$,  $p < .01$).\footnote{We use a Welch's instead of Student's
$t$-test since it does 
not assume that the distributions have equal variance.} Why is having
access to the correct morphological segmentation more advantageous for prefixed than suffixed complex
words? We argue that there are two key factors at play. First, the WordPiece tokenization sometimes generates the morphologically correct segmentation, but it does so with different frequencies for prefixes and  suffixes. 
To detect morphologically incorrect segmentations, we check whether the WordPiece segmentation keeps the stem intact,
which is in line with our definition of morphological validity  (Section \ref{sec:complex-nlp})
and provides a conservative estimate of the error rate.
For prefixes, the WordPiece tokenization is seldom correct (average error rate: $\mu = .903$, $\sigma = .042$), whereas for suffixes it is correct about half the time ($\mu = .503$, $\sigma = .213$). Hence, DelBERT gains a greater advantage for prefixed words. Second, prefixes and suffixes have different linguistic
properties that affect the prediction task in unequal ways. Specifically, 
whereas suffixes have both syntactic and semantic functions, prefixes 
have an exclusively semantic function and always add lexical-semantic meaning to the stem
\citep{Giraudo.2003, Beyersmann.2015}. As a result, cases such as \texttt{unamusing} where the affix
boundary
is a decisive factor for the prediction task are more likely 
to occur with prefixes than suffixes, thus increasing 
the importance of a morphologically correct segmentation.\footnote{Notice that there are suffixes with similar semantic effects (e.g., \texttt{less}), but they are less numerous.}

Given the differences between sentiment and topicality prediction, 
we expect variations in the relative importance of the two
identified factors:
(i)
in the case of sentiment the advantage of $s_d$ 
should be maximal for affixes directly affecting 
sentiment;
(ii)
in the case of topicality its advantage should be the larger
the higher the proportion of incorrect segmentations
for a particular affix, and hence the more frequent 
the cases where DelBERT has access to the stem while BERT does not. To test this hypothesis,
we focus on predictions for prefixed complex words. 
For each dataset, we measure for individual prefixes the accuracy increase of the DelBERT models
as compared to the BERT models, averaged across
random seeds, as well as the proportion of morphologically incorrect segmentations
produced by WordPiece. 
We then calculate linear regressions to predict the accuracy increases based on
the proportions of incorrect segmentations. This analysis shows
a significant positive correlation for ArXiv ($R^2 = .304$, $F(1, 41)= 17.92$, $p < 0.001$) and Reddit ($R^2 = .270$, $F(1, 40)= 14.80$, $p < 0.001$) but not for Amazon ($R^2 = .019$, $F(1, 41)= .80$, $p = .375$), which 
is in line with our expectations (Figure \ref{fig:results-top}). Furthermore, ranking 
the prefixes by accuracy increase for Amazon confirms that the most pronounced differences are found for prefixes that can change the sentiment such as \texttt{non}, \texttt{anti}, \texttt{mal}, and \texttt{pseudo} (Figure \ref{fig:results-sent}).

\subsection{Qualitative Analysis}

Besides quantitative factors, we are interested in identifying
qualitative 
contexts
 in which DelBERT has a particular advantage compared 
to BERT.
To do so, we filter the datasets for complex words that are consistently classified
correctly by DelBERT and incorrectly by BERT. Specifically, we compute for each word
the average likelihood of the true semantic class across DelBERT and BERT models, respectively, and rank words according to the likelihood difference between both model types.
Examining the words with the most extreme differences, we observe 
three classes (Table \ref{tab:qual-analysis}).

First, the addition of a suffix is often connected with
morpho-orthographic changes (e.g., the deletion of a
stem-final \texttt{e}), which leads to a segmentation of the stem into several subwords since the truncated stem is not in the WordPiece vocabulary (\texttt{applausive}, \texttt{isotopize}, \texttt{prematuration}). The model does not seem to be able 
to recover the meaning of the stem from the subwords. Second, the addition of a prefix
has the effect that the word-internal (as opposed to
word-initial) form of the stem
would have to be available for proper segmentation.
Since this
form rarely exists in the WordPiece vocabulary, the stem is segmented into several subwords
(\texttt{superannoying}, \texttt{antimicrosoft}, \texttt{nonmultiplayer}). Again, it does
not seem to be possible for the model to recover the meaning of the stem. Third, the segmentation
of prefixed complex words often fuses the prefix with the first characters of the 
stem (\texttt{overseasoned}, \texttt{inkinetic}, \texttt{promosque}). This case is particularly
detrimental since it not only makes it difficult to recover the meaning of the stem but also 
creates associations with unrelated meanings, sometimes even opposite meanings as in the case of \texttt{superbizarre}. The three classes thus underscore the difficulty of inferring the 
meaning of complex words from the subwords when the whole-word meaning
is not stored in the model weights and the subwords
are not morphological.

\section{Related Work}

Several recent studies have examined 
how the performance of PLMs is affected by their input segmentation. 
\citet{Tan.2020} show that tokenizing
inflected words into stems and inflection symbols allows BERT to generalize better on non-standard inflections. \citet{Bostrom.2020} pretrain RoBERTa with different 
tokenization methods and find tokenizations that align more
closely with morphology to perform better on a number of tasks. \citet{Ma.2020} show that providing BERT 
with character-level information also leads to enhanced performance.
Relatedly, studies from automatic speech recognition have
demonstrated that morphological decomposition improves the perplexity of language models \citep{Fang.2015, Jain.2020}.
Whereas these studies change the vocabulary of input tokens (e.g., 
by adding special tokens), we show that even when keeping the pretrained 
vocabulary fixed, employing it 
in a morphologically correct way leads to
better performance.\footnote{There 
are also studies that analyze morphological aspects of PLMs without a focus on questions surrounding segmentation \citep{Edmiston.2020, Klemen.2020}.}

Most NLP studies on derivational morphology have been devoted to the question of how semantic representations of derivationally complex words
can be enhanced by including morphological information \citep{Luong.2013, Botha.2014, Qiu.2014, Bhatia.2016, Cotterell.2018}, and how 
affix embeddings can be computed \citep{Lazaridou.2013, Kisselew.2015, Pado.2016}. \citet{Cotterell.2017}, \citet{Vylomova.2017}, and \citet{Deutsch.2018} propose sequence-to-sequence models for 
the generation of derivationally complex words. \citet{Hofmann.2020c} address the same task using BERT. 
In contrast, we analyze how different input segmentations affect the semantic representations of derivationally
complex words in PLMs, a question that has not been addressed before.

\section{Conclusion}

We have
examined how the input segmentation of PLMs, specifically BERT, 
affects their interpretations of derivationally complex words. Drawing upon 
insights from psycholinguistics, we have deduced a conceptual
interpretation of PLMs as serial dual-route models,
which implies that maximally meaningful input tokens should
allow for the best generalization on new words. This hypothesis
was confirmed by a series of semantic probing 
tasks on which DelBERT, a model using derivational segmentation,
consistently outperformed 
BERT using WordPiece segmentation. 
Quantitative and qualitative analyses further showed 
that BERT's inferior performance was caused by its inability
to infer the complex-word meaning as a function of the
subwords when the complex-word meaning was not stored
in the weights. Overall, our findings
suggest that the generalization capabilities of PLMs
could be further improved if a morphologically-informed vocabulary 
of input tokens were used.

\section*{Acknowledgements}

This work was funded by
the European Research Council (\#740516) and the Engineering and Physical Sciences Research Council
(EP/T023333/1). 
The first author was also supported by the German Academic Scholarship Foundation and the Arts and Humanities Research Council. 
We thank the reviewers for
their helpful comments.

\bibliography{acl2021}
\bibliographystyle{acl_natbib}

\appendix

\section{Appendices}

\subsection{Derivational Segmentation} \label{app:segmentation}

Let $A$ be a set of derivational affixes and $S$ a set of stems. To determine the 
derivational segmentation of a  
word $w$, we employ an iterative algorithm. 
Define the set $B^A_1$ of $w$ as the words
that remain when one derivational affix from $A$ is removed from $w$.  
For example, \texttt{unlockable} can be segmented into
\texttt{un}, \texttt{lockable} and \texttt{unlock}, \texttt{able}
so $B^A_1(\mbox{\texttt{unlockable}}) =
\{\mbox{\texttt{lockable}},\mbox{\texttt{unlock}}\}$ (we assume that \texttt{un} and \texttt{able} are in $A$).
We then iteratively create 
$B^A_{i+1}(w) = \bigcup_{b \in B^A_i(w)} B^A_1(b)$, i.e., we iteratively remove affixes from 
$w$. We stop as soon as 
$B^A_{i+1}(w) \cap  S  \neq \emptyset$. The element in this intersection,
together with the used affixes from $A$, forms the derivational
segmentation of $w$.\footnote{If $|B^A_{i+1}(w) \cap  S| > 1$ (rarely the case in practice), 
the element with the lowest number of suffixes is chosen.} If there is no $i$
such that $B^A_{i+1}(w) \cap  S  \neq \emptyset$, $w$ does not have a 
derivational segmentation.
The algorithm is
sensitive to most morpho-orthographic rules of English
\cite{Plag.2003}, e.g., when the suffix \texttt{ize} is removed from
\texttt{isotopize}, the resulting word is \texttt{isotope}, not
\texttt{isotop}.

In this paper, we follow \citet{Hofmann.2020c} in using BERT's prefixes,
suffixes, and stems as input to the algorithm.
Specifically, we assign 46 productive prefixes and 44 productive 
suffixes in BERT's vocabulary to $A$ and
all fully alphabetic words with more than 3 characters in BERT's vocabulary (excluding stopwords and affixes) to $S$, resulting in a total of 20,259 stems. This means that we only consider derivational segmentations
that are possible given BERT's vocabulary.

\subsection{Data Preprocessing} \label{app:preprocessing}

We exclude texts written in a language other than English and remove strings containing numbers as well as hyperlinks. We follow \citet{Han.2011} in reducing repetitions
of more than three letters (\texttt{niiiiice}) to three letters.

\subsection{Hyperparameters} \label{app:hyper}

The feed-forward network has a ReLU activation after the first layer
and a sigmoid activation after the second layer. The first layer has 100 dimensions. We apply dropout of 0.2 after the first layer. All other hyperparameters are as for BERT\textsubscript{BASE} (uncased) \citep{Devlin.2019}.
The number of trainable parameters is 109,559,241.

We use a batch size of 64 and perform grid search for the number of epochs $n \in \{1, \dots ,20\}$ and the learning rate $l \in \{ \num{1e-6}, \num{3e-6} , \num{1e-5}, \num{3e-5} \}$ (selection criterion: F1 score). We tune $l$ on Reddit (80 hyperparameter search trials per model type) and use the best configuration (which is identical for all model types) 
for 20 training runs with different random seeds on all three datasets (20 hyperparameter search trials per model type, dataset, and random seed). Models are trained with binary cross-entropy as the loss function and Adam \citep{Kingma.2015} as the optimizer. Experiments are performed on a GeForce GTX 1080 Ti GPU (11GB).

Table \ref{tab:hyper-stats} lists statistics of the validation performance
over hyperparameter search trials and provides
information about best hyperparameter configurations as well as runtimes.\footnote{Since expected validation performance \citep{Dodge.2019} may not be correct for grid search, we report mean and standard deviation of the performance instead.} See also Section \ref{sec:quant-analysis} and particularly Figure \ref{fig:convergence} in the main text, where we present a detailed analysis of 
the convergence behavior of the two main model types examined in this study (DelBERT and BERT).

\begin{table*}[b]\centering
\resizebox{\linewidth}{!}{%
\begin{tabular}{@{}lrrrrrrrrrrrrrrr@{}}
\toprule
{} & \multicolumn{5}{c}{Amazon}  & \multicolumn{5}{c}{ ArXiv} & \multicolumn{5}{c}{Reddit}\\
\cmidrule(lr){2-6}
\cmidrule(lr){7-11}
\cmidrule(l){12-16}
Model & $\mu$ & $\sigma$ & $n$ & $l$  & $\tau$ & $\mu$ & $\sigma$ & $n$ & $l$ & $\tau$ & $\mu$ & $\sigma$ & $n$  & $l$ & $\tau$\\
\midrule
DelBERT   & .627 & .007 & 6.75 & 3e-06 & 67.73 & .725 & .006 & 11.45 & 3e-06  & 28.69 & .687 & .006 & 5.45  &  3e-06 & 25.56 \\
BERT  & .612 & .006 & 7.30 & 3e-06 & 66.18 & .693 & .015 & 17.05& 3e-06  &28.04  & .657 & .007 & 9.25  &  3e-06 & 25.06  \\ 
\midrule
Stem   & .556 & .016 & 9.85 & 3e-06 &67.43 & .699 & .005 & 8.15 & 3e-06  & 28.56 & .670 & .006 & 6.00  &  3e-06 & 25.39 \\ 
Affixes   & .519 & .008 & 5.55 & 3e-06 & 67.70 & .599 & .004 & 7.50 & 3e-06  & 28.43 & .593 & .003 & 9.35  &  3e-06 & 25.49 \\ 
\bottomrule
\end{tabular}
}
\caption{Validation performance statistics and hyperparameter search details. The table shows the mean ($\mu$) and standard deviation ($\sigma$) of the validation performance (F1) on all hyperparameter search trials, the number of epochs ($n$) and learning rate ($l$) with the best
validation performance, and the runtime ($\tau$) in minutes for one full hyperparameter search (20 trials). The numbers are averaged across 20 training runs with different random seeds.}  \label{tab:hyper-stats}
\end{table*}

\end{document}